\documentclass[lettersize,journal]{IEEEtran}
\usepackage{amsmath,amsfonts}
\usepackage{algorithmic}
\usepackage{algorithm}
\usepackage{array}
\usepackage[caption=false,font=normalsize,labelfont=sf,textfont=sf]{subfig}
\usepackage{textcomp}
\usepackage{stfloats}
\usepackage{url}
\usepackage{verbatim}
\usepackage{graphicx}
\usepackage{cite}
\begin{document}
% from https://journals.ieeeauthorcenter.ieee.org/create-your-ieee-journal-article/authoring-tools-and-templates/

\title{Semantic Encoder Guided Generative Adversarial Face Ultra-Resolution Network}

\author{Xiang Wang,
        Yimin Yang,~\IEEEmembership{Senior Member, IEEE,} Qixiang Pang, Xiao Lu, and Yu Liu,~\IEEEmembership{Member,~IEEE, Shan Du,~\IEEEmembership{Senior Member, IEEE}}
        % <-this % stops a space
\thanks{X. Wang is with the Dept. of Computer Science, Lakehead University, Thunder Bay, ON, Canada, e-mail: xwang150@lakeheadu.ca}% <-this % stops a space
\thanks{Y. Yang is with the Dept. of electrical and computer engineering, Western University, London, ON, Canada, e-mail: yyan2294@uwo.ca}

\thanks{Q. Pang is with the Okanagan College, Kelowna, BC, Canada, e-mail: kpang@okanagan.bc.ca}

\thanks{X. Lu is with the College of Engineering and Design, Hunan Normal University, Hunan, China, e-mail: xlu\_hnu@163.com}

\thanks{Y. Liu is with the School of Microelectronics, Tianjin University, Tianjin, China, e-mail: liuyu@tju.edu.cn}

\thanks{S. Du is with the Dept. of Comp. Sci., Math, Physics \& Statistics, The University of British Columbia Okanagan, BC, Canada, e-mail: shan.du@ubc.ca}% <-this % stops a space
}
\markboth{Journal of \LaTeX\ Class Files,~Vol.~14, No.~8, August~2021}%
{Shell \MakeLowercase{\textit{et al.}}: A Sample Article Using IEEEtran.cls for IEEE Journals}


\maketitle

\begin{abstract}
Face super-resolution is a domain-specific image super-resolution, which aims to generate High-Resolution (HR) face images from their Low-Resolution (LR) counterparts. In this paper, we propose a novel face super-resolution method, namely Semantic Encoder guided Generative Adversarial Face Ultra-Resolution Network (SEGA-FURN) to ultra-resolve an unaligned tiny LR face image to its HR counterpart with multiple ultra-upscaling factors (e.g., 4× and 8×). The proposed network is composed of a novel semantic encoder that has the ability to capture the embedded semantics to guide adversarial learning and a novel generator that uses a hierarchical architecture named Residual in Internal Dense Block (RIDB). Moreover, we propose a joint discriminator which discriminates both image data and embedded semantics. The joint discriminator learns the joint probability distribution of the image space and latent space. We also use a Relativistic average Least Squares loss (RaLS) as the adversarial loss to alleviate the gradient vanishing problem and enhance the stability of the training procedure. Extensive experiments on large face datasets have proved that the proposed method can achieve superior super-resolution results and significantly outperform other state-of-the-art methods in both qualitative and quantitative comparisons.
\end{abstract}


\begin{figure*}[t!]
\centerline{\includegraphics[width =\textwidth]{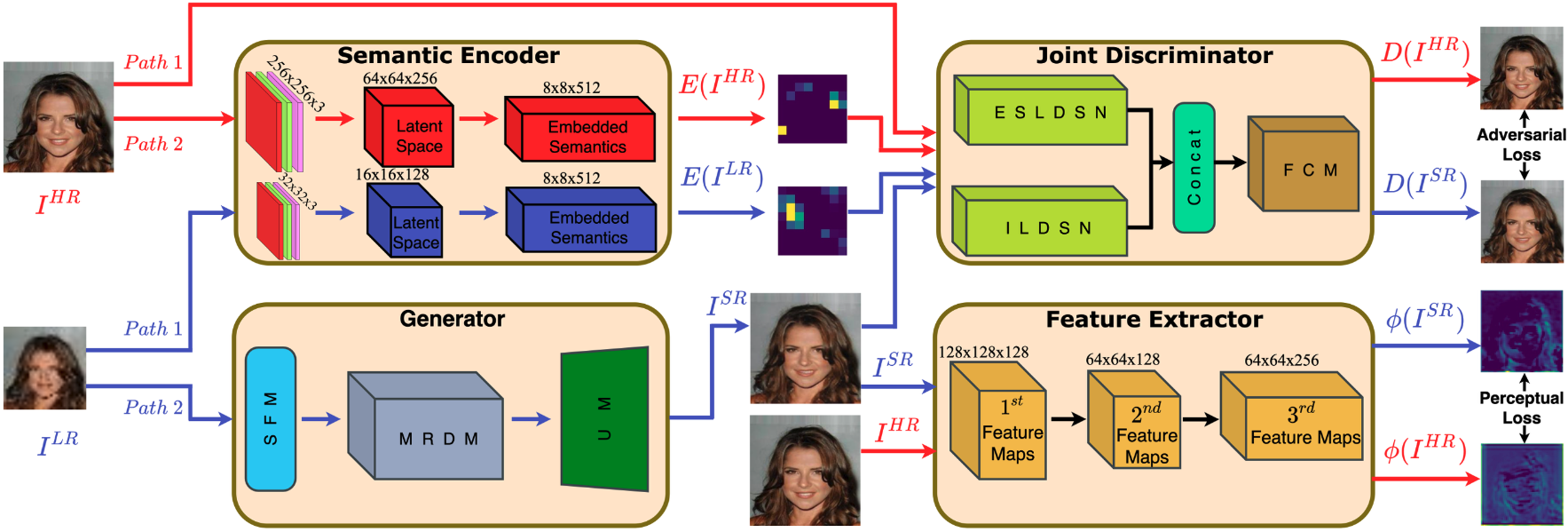}}
\caption{\textbf{Proposed SEGA-FURN and its components: Semantic Encoder $E$, Generator $G$, Joint Discriminator $D$ and Feature Extractor $\phi$.} For $D$, ESLDSN represents the Embedded Semantics-Level Discriminative Sub-Net, ILDSN represents the Image-Level Discriminative Sub-Net, and FCM denotes the Fully Connected Module. As for the generator $G$, there are three stages: Shallow Feature Module (SFM), Multi-level Residual Dense Module (MRDM), and Upsampling Module (UM). $I^{HR}$ and $I^{LR}$ denote HR face images and LR face images respectively. $I^{SR}$ is SR images from $G$. Furthermore, $E(\cdot)$ denotes the embedded semantics obtained from $E$. $D(\cdot)$ represents the output probability of $D$. $\phi(I^{HR})$ and $\phi(I^{SR})$ describe the features learned by $\phi$.}
\end{figure*}

\section{Introduction}
\IEEEPARstart{F}{ace} Super-Resolution (FSR) has been a promising computer vision topic in recent years. It is widely applied to the face applications such as face surveillance \cite{wilman_face_recoginition_problem}, identification \cite{face_recognition2}, and medical diagnosis \cite{isaac2015super}. However, there are many particular challenges in the research, such as when faces have complex facial expressions and when most of the facial information has been lost. To address these problems, many FSR methods have been proposed. 

The deep learning-based SR methods enhance the effect of super-resolution \cite{SRCNN,VDSR,Zhou,ma2020deep,hu2020face}. The first deep learning-based SR end-to-end manner model, SRCNN, was proposed by Dong $et$ $al.$ \cite{SRCNN}. To enhance the super-resolution performance, Zhu $et$ $al.$ \cite{Zhou} presented the deep cascaded bi-network incorporating face hallucination and dense correspondence field estimation. Recently, some methods resort to applying facial prior knowledge (e.g., facial landmarks, component maps, etc.) to super resolution process. By incorporating the face recovery model and landmark estimation model, Ma $et\ al.$ \cite{ma2020deep} used prior landmark maps to explore facial key structures. In \cite{hu2020face}, Hu $et\ al.$ proposed a 3D facial prior guided FSR method, where it utilizes 3D facial prior to estimating the facial spatial information. However, a large amount of prior knowledge is required in these prior knowledge based methods, which can cost huge computational resources and increase the time of the training procedure. Thus, they are not suitable for real-time applications.

% The HiFaceGAN \cite{yang2020hifacegan} proposed by Yang $et\ al.$ is trained with the multi-scale discriminator and the autoencoder to recover facial details.
Another common issue that occurred in the above-mentioned methods is that because of utilizing reconstruction objective loss $L_{1}$ or $L_{2}$ in the learning process, the super-resolved image tends to be overly smooth and incompatible with the HR image perceptually. To address the existing problem, Generative Adversarial Network (GAN) -based FSR methods have been investigated \cite{SRGAN,ESRGAN_Wang,FSRGAN,URDGN,TDAE, yu2020hallucinating,kim2019progressive}. It is proved that by introducing perceptual loss in the optimization process, these methods are capable of producing a visually plausible super-resolved image that is faithful to the HR version. As pioneer GAN-based SR methods, SRGAN \cite{SRGAN}, ESRGAN \cite{ESRGAN_Wang} composed of a generator and discriminator incorporates perceptual loss with adversarial loss to produce realistic images. Yu $et$ $al.$ involved GAN manner into FSR area, and proposed \cite{URDGN}. Furthermore, based on URDGN, Yu $et\ al.$ \cite{yu2020hallucinating} proposed a multiscale network, namely MTDN. Nevertheless, MTDN cannot handle the situation when the magnification factor is increased largely. The performance degraded greatly and severe distortions appeared in the results. FSRGAN \cite{FSRGAN} and FAN \cite{kim2019progressive} introduced geometry prior knowledge and landmark heatmaps of face images to the super resolution process. As aforementioned, these prior knowledge-based methods have computational burdens in the training procedure.

Besides those drawbacks, the aforementioned methods also severely suffer from the problem of model collapse and training instability, leading to notorious oversmoothing artifacts \cite{SRGAN,ESRGAN_Wang}. To break through these limitations and produce photo-realistic super-resolved face images, we propose a novel GAN-based SR method, namely Semantic Encoder guided Generative Adversarial Face Ultra-Resolution Network (SEGA-FURN). The overview of the proposed SEGA-FURN is demonstrated in Fig. 1, where it contains four components: Semantic Encoder, Joint Discriminator, Generator, and Feature Extractor. The LR image and HR image, both have two sets of paths. The set of paths shown in red indicates a real tuple, and symmetrically the set of paths shown in blue represents a fake tuple. As for the red paths (real tuple), Path 1 shows that the HR image is sent to the joint discriminator directly and Path 2 shows the HR image is passed to the semantic encoder, where the semantics extraction is employed on the HR image to acquire its embedded semantics. And for the blue paths (fake tuple), the two paths mean two different flows of LR images. For Path 1, the LR image is passed to the semantic encoder to extract its embedded semantics. And for Path 2, the LR image is fed into the generator, where the residual in the internal dense block (RIDB) is put forward as the basic building unit for the generator to obtain the image features at different levels and the inside upsampling operation is able to super resolve the LR image to be the realistic HR image with fine details. There are two sets of paths entering into the joint discriminator. The red tuple includes the HR image and its embedded semantics. Similarly, the blue tuple is composed of the SR image produced by the generator and embedded semantics extracted from the LR image. The joint discriminator aims to evaluate how an input real tuple is more authentic compared to a given fake generated tuple. In addition, both SR and HR images are put into a feature extractor to extract the intermediate feature maps which are used to calculate the perceptual loss. Such a learning strategy allows our proposed method to achieve superior SR performance compared with other state-of-the-art methods. It is noted that the basic idea of SEGA-FURN is recorded by the first author's dissertation \cite{wang2021deep}.

The main contributions of our proposed method can be summarized as follows:

1) Our proposed method is able to ultra-resolve an unaligned tiny face image to a Super-Resolved (SR) face image with multiple upscaling factors (e.g., 4× and 8×). In addition, our method does not need any prior information or facial landmark points.

2) We design a semantic encoder to reverse the image information back to the embedded semantics reflecting facial semantic attributes. The embedded semantics combined with image data is fed into a joint discriminator. Such innovation enables the semantic encoder to guide the discriminative process, which enhances the discriminative ability of the discriminator.

3) We propose the Residual in Internal Dense Block (RIDB) as the basic architecture for the generator. This innovation provides an effective way to take advantage of hierarchical features, increasing the feature extraction capability of the SEGA-FURN. This architecture was partially described in our previous paper \cite{xiang2021end-to-end}

4) We propose a joint discriminator which is capable of learning the joint probability constructed by embedded semantics and visual information (HR and LR images), resulting in a powerful discriminative ability. Furthermore, in order to remedy the problem of vanishing gradient and improve the model stability, we make use of RaLS objective loss to optimize the training process.

The remainder of this paper is organized as follows. In Section 2, we provide some preliminary information on our proposed method to solve the problem. In Section 3, we provide a detailed description of the proposed method and its major components: generator, semantic encoder, joint discriminator, feature extractor, and loss functions. The novelty of the proposed method is discussed as well. In Section 4, a large amount of experimental results are presented and compared with other state-of-the-art methods. Both qualitative and quantitative analyses are conducted and performance superiority over other methods are demonstrated. Section 5 concludes this paper.

\begin{figure*}[t!]
%\centerline{\includegraphics[width =\textwidth]{ICASSP_RIDB_DNB.png}} 6.2 to 6/// 6.4-6.5
% \centerline{\includegraphics[width =\textwidth,height=6.5cm
\centerline{\includegraphics[width =\textwidth]{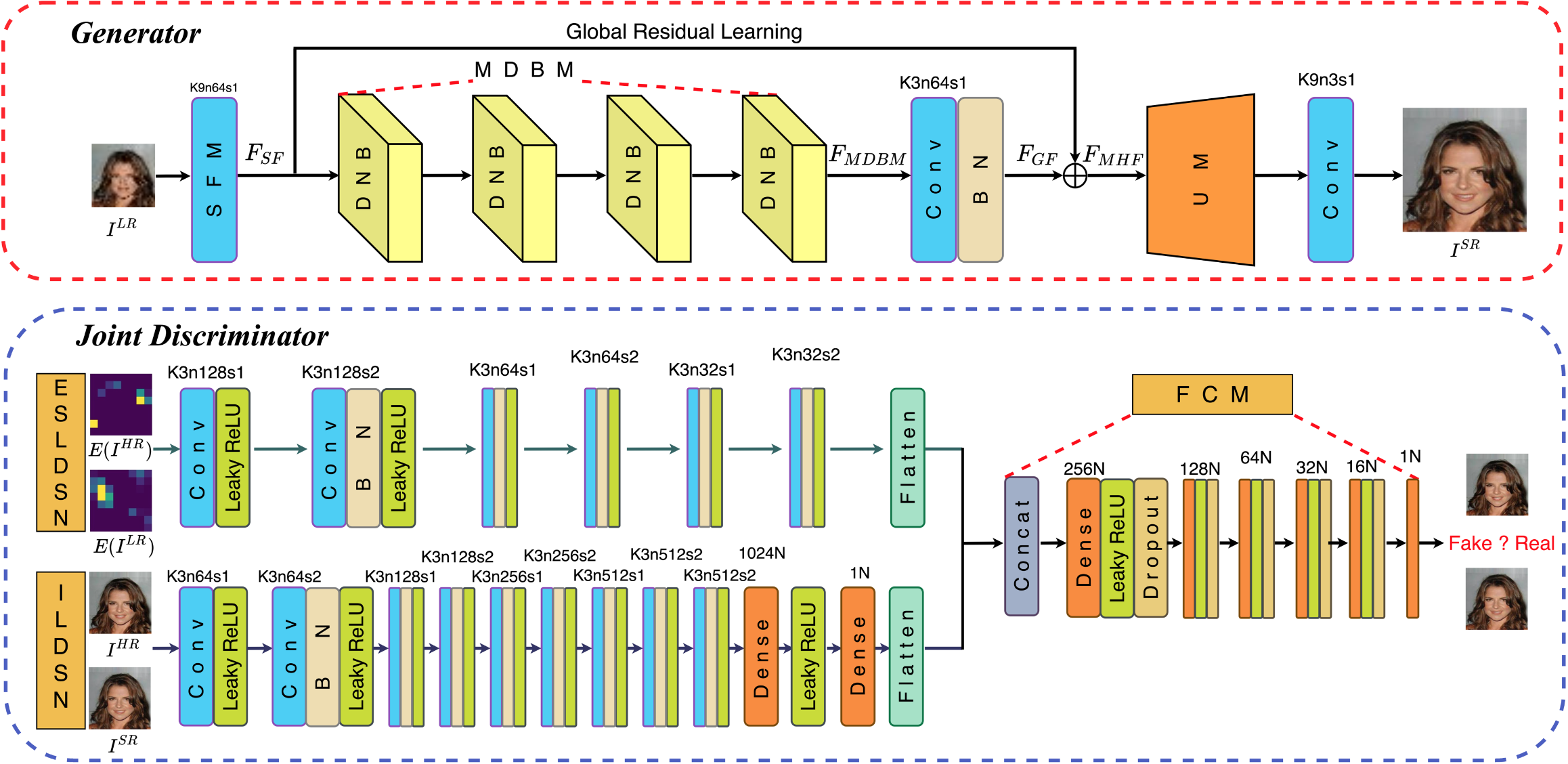}}
\caption{\textbf{Red dotted rectangle}: The architecture of the Generator. \textbf{Blue dotted rectangle}: The architecture of the Joint Discriminator. $F_{SF}$ denotes shallow features, $F_{MDBM}$ denotes the outputs of MDBM, $F_{GF}$ represents global features, and $F_{MHF}$ represents multiple hierarchical features. K, n, and s are the kernel size, number of filters, and strides respectively. N is the number of neurons in a dense layer.}
\label{fig}
\end{figure*}

\section{Preliminary}
In this section, we first review the Standard Generative Adversarial Network (SGAN) \cite{GAN}. Secondly, we analyze the major problems associated with SGAN and present improved GAN variants that can address current problems. Many GAN-based SR methods utilize SGAN adversarial objective function as their optimization strategy, but SGAN may cause unexpected issues resulting in unstable training procedures and poor super resolution results \cite{RaGAN,WGAN-GP,LSGAN}. In the following paragraphs, we analyze the underlying problems raised by involving GAN in the SR methods and then explore effective solutions, so as to improve super resolution performances. 

\subsection{Overview of SGAN}
SGAN consists of two networks, one of which is the Generator $G$, and the other is the discriminator $D$. SGAN has been applied in many applications, such as super-resolution \cite{SRGAN}, image translation \cite{zhu2017unpairedcycle_gan} and face aging \cite{faceaging_gans}. Through adversarial learning in the SGAN, the generator and discriminator compete against each other. Both two networks try to optimize themselves to solve the adversarial max-min problem. The objective function used in SGAN is:
\begin{equation}
\begin{split}
V(G,D) = & \max \limits_{d} \min \limits_{g}\mathbb{E}_{x\sim p_{data}(x)}[logD_{d}(x)] \\ 
& + \mathbb{E}_{z\sim p_{z}(z)}[log(1-D_{d}(G_{g}(z)))]
\end{split}
\end{equation}where $V(G, D)$ is a binary cross entropy loss which is commonly used in GAN applications, $G$ is supposed to map random noise $z$ from the prior distribution $p_{z}(z)$ over real data $x$, and $D$ is the discriminator which distinguishes whether its input comes from the $G$ or real data distribution $p_{data}(x_{r})$. The ultimate goal of SGAN is that $D$ and $G$ are capable of reaching the Nash equilibrium state, in which once SGAN attains Nash equilibrium, the generator can generate realistic-looking images which fool the discriminator.

\subsection{Problem Analysis}
However, SGAN in \cite{GAN,URDGN,SRGAN} encounters the problem of gradient vanishing, model collapse, and poor quality of the generated images. Several works \cite{LSGAN, RaGAN, WGAN} have proved that the objective function of the SGAN causes vanishing gradients, resulting in the instability of GAN training. The discriminator of the SGAN can be expressed as Eqn. 2:
\begin{equation}
    D(x) = \sigma(C(x))
\end{equation}where $x$ denotes either $I^{HR}$ or $I^{SR}$ in this context, $\sigma$ represents the sigmoid function, $C(x)$ is the probability predicted by the non-transformed discriminator. The restriction of SGAN is that they only concentrate on increasing the probability that fake samples belong to real rather than decreasing the probability that real samples belong to real simultaneously. In SGAN, if the optimal discriminator is reached, it will stop learning the real data but will only focus on the fake samples. As a result, the generator cannot receive enough gradient information from real data to make progress and the authenticity of fake samples will no longer be improved. To address this problem, several improved GAN variants have been proposed to find the objective function with smoother and non-vanishing gradients.

\subsection{Relativistic average GAN}
RaGAN \cite{RaGAN} was proposed as an improved SGAN by drawing up a Relativistic average Discriminator (RaD). The RaD demonstrates that the SGAN ignores the relative discriminant information between real samples and fake samples. This key property is complemented in RaD, which RaD not only improves the probability that the generated samples are real but also decreases the possibility that the real samples are real. The RaD can be expressed as:
\begin{equation}
    D(x_{r}, x_{f}) = \sigma (C(x_{r}) - E_{x_{f}}[C(x_{f})])
\end{equation}where $E_{x_{f}}$ denotes the average of the fake samples in one batch predicted by RaD. As Eqn. 3 shows, the RaD is capable of evaluating the probability that the real image is more realistic than the fake image, which addresses the issue of vanishing gradient and improves the stability of GAN.

\subsection{Least Squares GAN}
The Least Squares GAN (LSGAN) \cite{LSGAN} indicated that the vanishing gradient problem is mainly caused by the discriminator of SGAN using the sigmoid cross entropy loss. This work argued that the original discriminator penalizes a small decision error to update the generator which makes the generated samples stay on the correct side of the decision boundary, but are still far from corresponding real samples, leading to a vanishing gradient problem during the adversarial process. Motivated by this issue, LSGAN proposed the least squares loss function to penalize large errors coming from fake samples that lie far away from the decision boundary. The formulation can be expressed as:
\begin{equation}
\begin{aligned}
L_{D}^{LSGAN} = & \mathbb{E}_{x_{r}\sim p_{x_{r}}}[( C_{d}( x_{real})-0)^{2}] \\
& +\mathbb{E}_{x_{f}\sim p_{x_{f}}}[( C_{d}(x_{fake})-1)^{2}]
\end{aligned}
\end{equation}Thus, the discriminator utilizing least squares loss is capable of providing sufficient gradients when optimizing the generator, which is able to remedy the vanishing gradient problem. In our proposed method, we further take advantage of the Relativistic average Least Squares (RaLS) loss function which incorporates RaD into least squares loss. Such an optimization strategy can help our method stabilize the learning process and accelerate the convergence speed, producing authentic SR images.

\begin{figure*}[t!]
%\centerline{\includegraphics[width =\textwidth]{ICASSP_RIDB_DNB.png}}/// 4.32-4.5 height = 4.32cm
\centerline{\includegraphics[width =\textwidth]{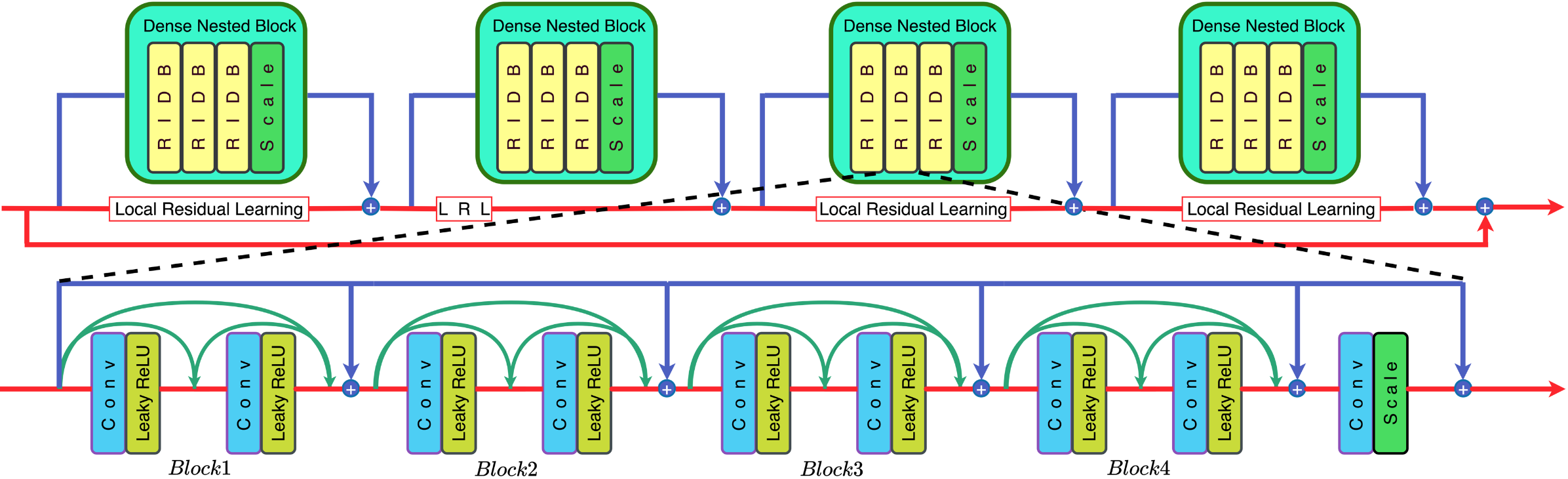}}
\caption{\textbf{Top}: Dense Nested Block (DNB) consists of multiple RIDBs. \textbf{Bottom}: The proposed Residual in Internal Dense Block (RIDB).}
\label{fig}
\end{figure*}

\section{Proposed Method}
In this section, we present the proposed method SEGA-FURN in detail. First, we describe the novel architecture of SEGA-FURN through four main parts: generator, semantic encoder, joint discriminator, and feature extractor. Next, we introduce the objective loss function RaLS \cite{RaGAN} for optimizing the generator and discriminator respectively. Furthermore, we discuss the benefit of the perceptual loss used in our method. Finally, we provide the overall total loss used in SEGA-FURN. The overview of our method is shown in Fig. 1. The architecture of the discriminator and generator can be seen in Fig. 2. Moreover, the structures of our proposed DNB and RIDB \cite{xiang2021end-to-end} are presented in Fig. 3.

\subsection{Generator}
We use the Dense Nested Blocks (DNBs) and Residual in Internal Dense Blocks (RIDBs) that we proposed in GAFH-RIDN \cite{xiang2021end-to-end} as basic units to construct $G$. As shown at the top of Fig. 2, the proposed generator mainly consists of three stages: Shallow Feature Module (SFM), Multi-level Dense Block Module (MDBM), and Upsampling Module (UM). The LR face image $I^{LR}$ is fed into the SFM as the initial input. In the end, SR face image $I^{SR}$ is obtained from the UM. The overall super-resolution process can be formulated as:
\begin{equation}
I^{SR}=G(I^{LR})
\end{equation}Specifically, the MDBM is built up by multiple DNBs formed by several RIDBs. As shown in Fig. 3, each DNB includes 3 RIDBs cascaded by residual connections and one scaling layer. The RIDB is able to extract hierarchical features and address the vanishing-gradient problem, which is the commonly encountered issue in \cite{SRGAN,ESRGAN_Wang,SRDenseNet,URDGN,RDN}. Owing to the designed architecture, the feature maps of each layer are propagated into all succeeding layers, promoting an effective way to extract hierarchical features and alleviating the gradient vanishing problem. It is emphasized that Local Residual Learning (LRL) is introduced to take effective use of the local residual features extracted by RIDBs. In addition, in order to help the generator fully take advantage of hierarchical features, we design the Global Residual Learning (GRL) to fuse the shallow features $F_{SF}$ and global features $F_{GF}$. Overall, benefiting from the proposed architecture \cite{xiang2021end-to-end}, the generator is capable of exploiting abundant hierarchical features and super-resolving from the LR space to the HR space.

\subsection{Semantic Encoder}
The proposed semantic encoder is designed to extract embedded semantics (as shown in Fig. 1), which is used to project visual information (HR, LR) back to the latent space. The GAN-based SR models \cite{SRGAN,ESRGAN_Wang,URDGN} only exploit visual information during the discriminative procedure, ignoring the semantic information reflected by latent representation. The proposed semantic encoder will complement the missing critical property. Previous GAN's work \cite{BiGAN,ALI} has proved that the semantic representation is beneficial to the discriminator.

Based on this observation, the proposed semantic encoder is designed to inversely map the image to the embedded semantics. The most significant advantage of the new semantic encode is that it can guide the discriminative process since the embedded semantics obtained from the semantic encoder can reflect semantic attributes, such as the facial features (shape and gender) and the spatial relationship between various components of the face (eyes, mouth). According to the methodology of semantic encoder and ablation studies, this advance is verified. It can be emphasized that the embedded semantics is fed into the joint discriminator along with HR and LR images. Thanks to this property, the semantic encoder can guide the discriminator to optimize, thereby enhancing its discriminative ability.

In this context, we use two side-by-side pre-trained VGG19 \cite{VGG19} networks as the semantic encoder to obtain the embedded semantics of the HR face image and the LR face image from different convolutional layers respectively. These two side-by-side VGG19 networks have the same structure except for different input dimensions since it needs to satisfy the different dimensions of the HR and LR face image respectively. 

% The dimension of both two embedded semantics is 8×8×512.

\begin{figure*}[t!]
 \centerline{\includegraphics[width =\textwidth]{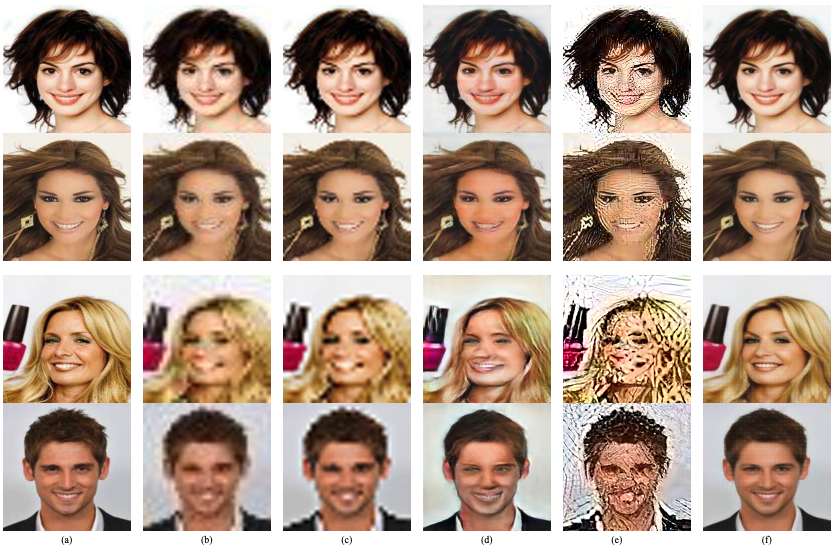}}
\caption{Qualitative comparison against state-of-the-art methods. \textbf{The top two rows}: the results of 4× upscaling factor from 32×32 pixels to 256×256 pixels. \textbf{The bottom two rows}: the results of 8× upsampling factor from 16×16 pixels to 256×256 pixels. From left to right: (a) HR images, (b) LR inputs, (c) Bicubic interpolation, (d) Results of SRGAN \cite{SRGAN}, (e) Results of ESRGAN \cite{ESRGAN_Wang}, and \textbf{(f) Our method.}}
\label{fig}
\end{figure*}

\subsection{Joint Discriminator}
As shown in Fig. 1, the proposed joint discriminator takes the tuple incorporating both visual information and embedded semantics as the input, where Embedded Semantics-Level Discriminative Sub-Net (ESLDSN) receives the input embedded semantics while the image information is sent to Image-Level Discriminative Sub-Net (ILDSN). Next, through the operation of the Fully Connected Module (FCM) on a concatenated vector, the final probability is predicted. Thus, the joint discriminator has the ability to learn the joint probability distribution of image data ($I^{HR}, I^{SR}$) and embedded semantics ($E(I^{HR}), E(I^{LR})$). There are two sets of paths entering into the joint discriminator. The set of paths shown in red indicates the real tuple which consists of the real sample $I^{HR}$ from the dataset and its embedded semantics $E(I^{HR})$. For the blue paths, a fake tuple is constructed from SR image $I^{SR}$ generated from the generator and $E(I^{LR})$ obtained from the LR image through a semantic encoder. As a result, different from \cite{SRGAN,ESRGAN_Wang,URDGN,FSRGAN}, our joint discriminator has the ability to evaluate the difference between real tuple $(I^{HR},E(I^{HR}))$ and fake tuple $(I^{SR},E(I^{LR}))$.

Moreover, in order to alleviate the problem of gradient vanishing and enhance the model stability, we adopt the Relativistic average Least Squares GAN (RaLSGAN) objective loss for the joint discriminator by applying the RaD to the least squares loss function \cite{LSGAN}. Let's denote the real tuple by $X_{real}=(I^{HR},E(I^{HR}))$ and denote the fake tuple by $X_{fake} = (I^{SR},E(I^{LR}))$. The process that makes joint discriminator relativistic can be expressed as follows:
\begin{equation}
\begin{aligned}
\tilde{C}(X_{real}) = (C(X_{real}) - E_{x_{f}}[C(X_{fake})]) \\
\tilde{C}(X_{fake}) = (C(X_{fake}) - E_{x_{r}}[C(X_{real})])
\end{aligned}
\end{equation}where $\tilde C(\cdot)$ denotes the probability predicted by joint discriminator, $E_{x_{f}}$ and $E_{x_{r}}$ describe the average of the SR images (fake) and HR images (real) in a training batch. Moreover, the least squares loss is used to measure the distance between HR and SR images. According to Eqn. 7, we optimize the joint discriminator by adversarial loss $L_{D}^{RaLS}$ and the generator is updated by $L_{G}^{RaLS}$, as in Eqn. 8.

\begin{equation}
\begin{aligned}
L_{D}^{RaLS}= & \mathbb{E}_{I^{HR}\sim p_{(I^{HR})}}[( \tilde{C}( X_{real})-1)^{2}] \\
& +\mathbb{E}_{I^{SR}\sim p_{(I^{SR})}}[( \tilde{C}( X_{fake})+1)^{2}]
\end{aligned}
\end{equation}
\begin{equation}
\begin{aligned}
L_{G}^{RaLS}= & \mathbb{E}_{I^{SR}\sim p_{(I^{SR})}}[( \tilde{C}( X_{fake})-1)^{2}] \\
& +\mathbb{E}_{I^{HR}\sim p_{(I^{HR})}}[( \tilde{C}( X_{real})+1)^{2}]
\end{aligned}
\end{equation}where $I_{HR} \sim P_{I^{HR}}$ and $I_{SR} \sim P_{I^{SR}}$ indicate the HR images and SR images distribution respectively. In addition, with the help of least squares loss and relativism in RaLS, SEGA-FURN is remarkably more stable and generates authentic and visually pleasant SR images.

In a joint discriminator, the ESLDSN takes the embedded semantics as the input and downsamples it through 6 convolutional layers with 3×3 kernels and the stride of 1 or 2 alternately, and then reshapes it to a 32-dimensional vector. The ILDSN receives an image and performs feature extraction through 9 groups of convolutional layers using the kernel of the size 3×3 followed by the LeakyReLU \cite{LeakyReLU} and the Batch Normalization (BN) layer to obtain the flattened 64-dimensional vector. Next, the resulting two vectors are concatenated by the concatenation layer and then fed into the FCM. As for FCM, it contains six dense layer blocks, where each dense layer block includes a dense layer, a LeakyReLU activation layer, and a dropout layer except for the last single dense layer. These six dense layers have 256, 128, 64, 32, 16, and 1 neurons respectively. Finally, the output is the probability that how the given HR face image is more realistic than the Super-Resolved (SR) face image.

\subsection{Feature Extractor} 
We further exploit pre-trained VGG19 \cite{VGG19} network as feature extractor $\phi$ in SEGA-FURN to obtain feature representations used to calculate the perceptual loss $L_{perceptual}$, where $L_{perceptual}$ is utilized in SEGA-FURN to eliminate the facial ambiguity and recover missing details of SR images. Instead of using high-level features as in SRGAN, ESRGAN for perceptual loss, we adopt low-level features before the activation layer (i.e., feature representations from `Conv3\_3' layer in the feature extractor), which contains complex edge texture. 

\subsection{Loss Function}
We involve perceptual loss $L_{perceptual}$ to constrain the intensity and feature similarities between HR and SR images \cite{Content_loss_merit1,content_loss_merit2}. Furthermore, adversarial loss $L_{G}^{RaLS}$ is adopted to super-resolve SR images containing visually appealing details and faithful to the HR.
\par
\textbf{Perceptual Loss}:
$L_{perceptual}$ is able to reduce the gap between the SR image and the HR image. It is formulated as:
\begin{equation}
L_{perceptual}=\frac{1}{WH}\sum_{q=1}^{W}\sum_{r=1}^{H}( \phi_{i,j}(I^{HR})_{q,r}-\phi_{i,j}( I^{SR})_{q,r})^{2}
\end{equation}where $W$, $H$ describe the height and width of the feature maps, $\phi(\cdot)$ denotes the output of feature extractor, $\phi_{i,j}$ indicates the feature representations obtained from $j$-th convolution layer before $i$-th maxpooling layer. 
\par
\textbf{Total Loss}:
The total loss function $L_{perceptual}$ for the generator can be represented as a weighted combination of two parts: perceptual loss $L_{perceptual}$ and adversarial loss $L_{G}^{RaLS}$, the formula is described as follows:
\begin{equation}
 L_{total} = \lambda _{con}L_{perceptual} + \lambda_{adv}L_{G}^{RaLS}
\end{equation}where $\lambda_{con}$, $\lambda_{adv}$ are the trade-off weights for the $L_{perceptual}$ and the $L_{G}^{RaLS}$. We set $\lambda_{con}$, $\lambda_{adv}$ empirically to 1 and $10^{-3}$ respectively.

\begin{figure*}[t!]
%\centerline{\includegraphics[scale=0.3]{ICASSP_celeba_8x.png}} ,height=8.4cm
 \centerline{\includegraphics[width =\textwidth]{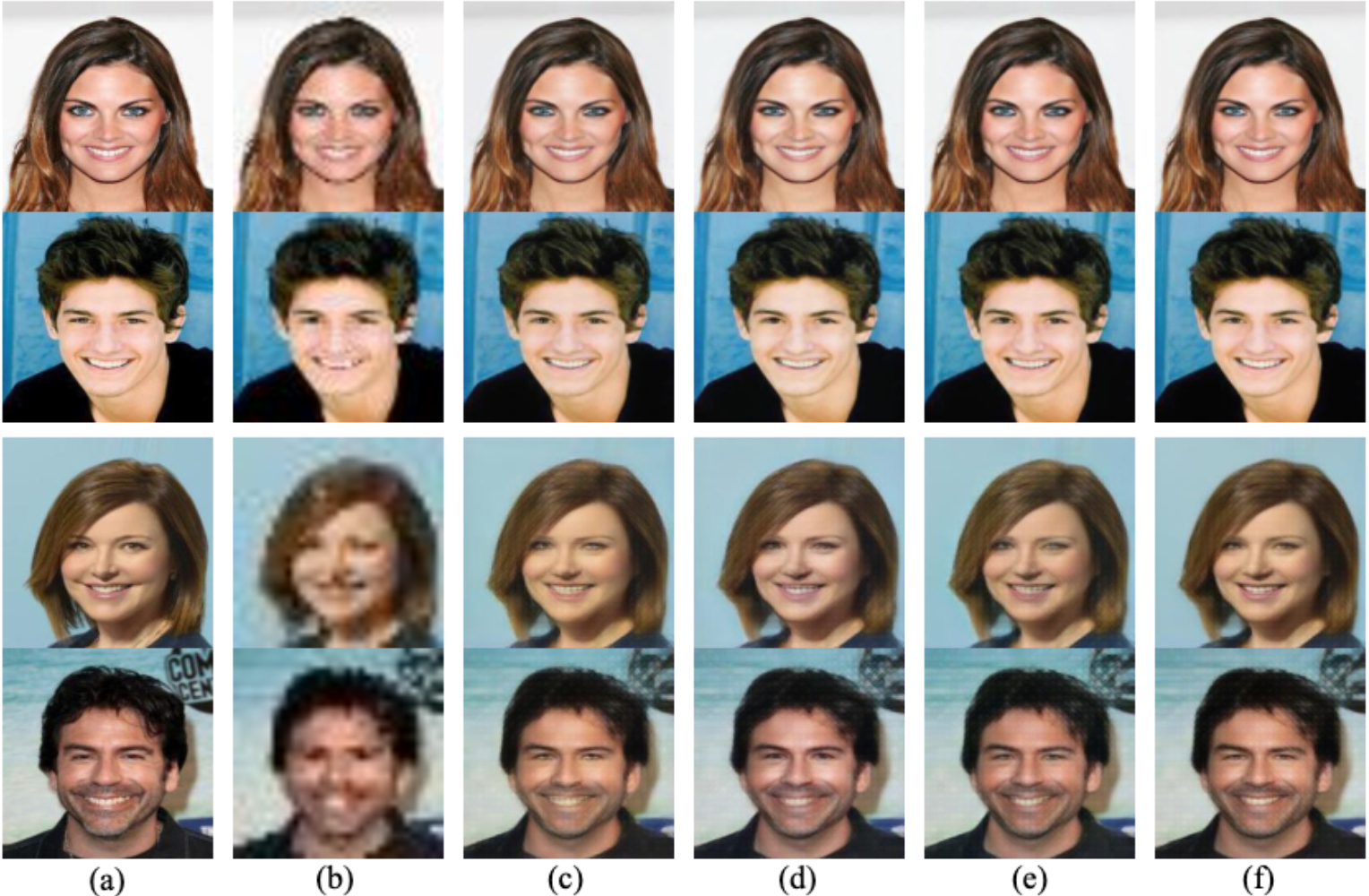}}
\caption{Qualitative comparison of ablation studies. \textbf{The top two rows:} the results of upscaling factor 4×. \textbf{The bottom two rows:} the results of upscaling factor 8×. From left to right: (a) HR images, (b) LR inputs, (c) Results of RIDB-Net, (d) Results of RIDB-RaLS-Net (e) Results of RIDB-SE-Net, and \textbf{(f) Results of RIDB-SE-RaLS-Net (SEGA-FURN).}}
\label{fig}
\end{figure*}

\section{Experiments}
In this section, we first present the details of the dataset and training implementation. Then, we demonstrate the experiments and evaluation results. We further compare our method with state-of-the-art methods. Moreover, in order to prove the effectiveness of SEGA-FURN, we conduct ablation experiments to verify the contributions of the proposed components.

Please note that we involved experimental results which were demonstrated in the corresponding published papers. In addition, the quantitative results of some state-of-the-art methods were missing when the upscaling factor is 4×, because they did not conduct experiments for upscaling factor 4×, and some methods did not exploit Structural Similarity (SSIM) as the evaluation criterion. For the qualitative comparison, we adopted the traditional super-resolution method, Bicubic interpolation, and the Generative Adversarial Network (GAN) \cite{GAN} -based super-resolution methods, SRGAN \cite{SRGAN} and ESRGAN \cite{ESRGAN_Wang}. Note that, in order to compare the visual effects fairly, we used their official published code of the above methods. As for SRGAN and ESRGAN, we download their models and scripts from the official GitHub repository$^{1,2}$\footnotetext[1]{https://github.com/tensorlayer/srgan}\footnotetext[2]{https://github.com/xinntao/ESRGAN} respectively. Moreover, the results of ESRGAN have been confirmed by the author of ESRGAN. 

\subsection{Datasets}
We conducted experiments on the public large-scale CelebFaces Attributes dataset, CelebA \cite{CelebA}. It consists of 200K celebrity face images of 10,177 celebrities. We used a total of 202,599 images, where we randomly selected 162,048 HR face images as the training set, and all the rest 40,511 images were chosen as the testing set.

\subsection{Implementation Details}
To verify the effectiveness of our method, we conducted experiments with multiple upscaling factors 4× and 8× respectively. We resized and cropped the images to 256x256 pixels as our HR face images do not have any alignment operation. In order to obtain two groups of LR downsampled face images, we used the bicubic interpolation method with downsampling factor $r$=4 to produce 64×64 pixels, and factor $r$=8 to produce 32×32 pixels LR images.

We trained our network 30k epochs using the Adam optimizer \cite{Adam} by setting $\beta _{1}$=0.9, $\beta _{2} $=0.999 with the learning rate of $10^{-4}$ and batch size of 8. We alternately updated the generator and discriminator until the model converges. For the quantitative comparison, we adopted Peak Signal-to-Noise Ratio (PSNR) and Structural Similarity (SSIM) as the evaluation metrics.

\subsection{Comparisons with State-of-the-art Methods}
We compare the proposed method with the state-of-the-art SR methods \cite{Yang21.35,SRCNN,Ma,CSGM,DIP,yang_23.07_yang2013structured,CBN,URDGN,IAGAN,TDN,FSRFCH,SRGAN,ESRGAN_Wang,TDAE,FaceAttr,VDSR,Liu} quantitatively and qualitatively. 
\label{ssec:subhead}

\begin{table}[t!]
\caption{Quantitative comparison on CelebA dataset for upscaling factor 4× and 8×, in terms of average PSNR(dB) and SSIM. Numbers in bold are the best evaluation results among state-of-the-art methods.}
\small
% \begin{center}
% \centering
\setlength{\tabcolsep}{2.5mm}
\begin{tabular}{lccllc}
\hline
\hline
Method & \multicolumn{2}{c}{\begin{tabular}[c]{@{}c@{}}CelebA 4×\\ PSNR \quad \ \ \ SSIM\end{tabular}} & \multicolumn{1}{c}{} & \multicolumn{2}{c}{\begin{tabular}[c]{@{}c@{}}CelebA 8×\\ PSNR \quad \ \ \ SSIM\end{tabular}} \\ \hline
Bicubic & 26.50 & 0.79 &  & 22.90 & 0.65 \\
Yang $et\ al.$ \cite{Yang21.35} & - & - &  & 21.35 & 0.60 \\
SRCNN \cite{SRCNN} & 28.93 & 0.79 &  & 23.11 & 0.65 \\
Ma $et\ al.$ \cite{Ma} & - & - &  & 23.12 & 0.64 \\
FAN \cite{kim2019progressive} & - & - &  & 22.96 & 0.69 \\
CSGM-BP \cite{CSGM} & 26.44 & - &  & 22.71 & - \\
DIP \cite{DIP} & 27.35 & - &  & 23.45 & - \\
IAGAN \cite{IAGAN} & 27.16 & - &  & 23.49 & - \\
Yang $et\ al.$ \cite{yang_23.07_yang2013structured} & - & - &  & 23.07 & 0.65 \\
FSRNet \cite{FSRGAN} &  &  &  & 21.19 & 0.60 \\
FSRGAN \cite{FSRGAN} &  &  &  & 20.95 & 0.51 \\
CBN \cite{CBN} & 29.37 & 0.79 &  & 18.77 & 0.54 \\
URDGN \cite{URDGN} & 29.10 & 0.79 &  & 24.82 & 0.70 \\
TDN \cite{TDN} & - & - &  & 22.66 & 0.66 \\
FSRFCH \cite{FSRFCH} & - & - &  & 23.14 & 0.82 \\
SRGAN \cite{SRGAN} & 26.76 & 0.82 &  & 20.64 & 0.62 \\
ESRGAN \cite{ESRGAN_Wang} & 23.82 & 0.71 &  & 20.32 & 0.57 \\
TDAE \cite{TDAE} & 29.49 & 0.81 &  & 20.40 & 0.57 \\
FaceAttr \cite{FaceAttr} & 29.78 & 0.82 &  & 21.82 & 0.62 \\
WGAN-GP \cite{WGAN-GP} & 23.28 & 0.69 & \multicolumn{1}{c}{} & - & - \\
VDSR \cite{VDSR} & - & - &  & 19.58 & 0.57 \\
Liu $et\ al.$ \cite{Liu} & - & - & \multicolumn{1}{c}{} & 21.60 & 0.66 \\
Ours & \textbf{30.14} & \textbf{0.87} &  & \textbf{25.21} & \textbf{0.73} \\ \hline \hline
\end{tabular}
%\end{center}
\end{table}
\par
\textbf{Qualitative Comparison}: The 4× and 8× qualitative results are depicted in Fig. 4. The top two rows show the 4× visual results. As for Bicubic interpolation, we observe that its results contain over-smooth visualization effects. SRGAN \cite{SRGAN} relatively enhances the SR results compared to Bicubic interpolation, but it still fails to generate fine details, especially in facial components, such as eyes, and mouth. It is obvious that ESRGAN \cite{ESRGAN_Wang} produces overly smoothed visual results and misses specific textures. On the contrary, the 4× SR images produced by our method retain facial-specific details and are faithful to HR counterparts.

To reveal the powerful super-resolution ability of our proposed method, we further conduct experiments with 8× ultra upscaling factor. As shown in the bottom two rows in Fig. 4, the SR visual quality obtained by Bicubic interpolation, SRGAN, and ESRGAN is decreased, since the magnification is increased and the correspondence between HR and LR images becomes incompatible. The outputs of Bicubic interpolation generate unpleasant noises. SRGAN encounters a mode collapse problem during the super-resolution process so it produces severe distortions in SR images. As for ESRGAN, it produces SR images that show broken textures and noticeable artifacts around facial components. The other methods all have their respective problems with the 8× ultra scaling factor. In contrast, our method is capable of producing photo-realistic SR images which preserve perceptually sharper edges and fine facial textures. 
\par
\textbf{Quantitative Comparison}: The quantitative results with multiple ultra upscaling factors 4× and 8× are shown in Table I. It is obvious that our method attains the best in both PSNR and SSIM evaluations, 30.14dB/0.87 for 4× and 25.21dB/0.73 for 8×, among all methods. FaceAttr \cite{FaceAttr} is the second best method for 4×, 29.78dB/0.82, however, it degrades dramatically and performs poorly when the upscaling factor increases to 8×, obtaining 21.82dB/0.62. In contrast, our method ranks the first for both upscaling factors 4× and 8×, which reflects the robustness of the proposed SEGA-FURN. Moreover, it is notable that our proposed method not only boosts PSNR/SSIM by a large margin of 1.04dB/0.08 over the classic method URDGN \cite{URDGN} with upscaling factor 4× but also is higher than URDGN which is the second best for the 8× upscaling. This result proves the stability of our method with multiple upscaling factors. In addition, we compare with SRGAN \cite{SRGAN} and ESRGAN \cite{ESRGAN_Wang} which also use generative adversarial structure. It is obvious that our method not only improves SR image quality from a perceptual aspect but also achieves impressively numerical results.
% we provide the quantitative results with multiple ultra upscaling factors 4× and 8×.
\begin{table}[t!]
\caption{Description of SEGA-FURN variants with different components in experiments.}
\small
% \begin{center}
% \centering
\setlength{\tabcolsep}{2mm}
\begin{tabular}{lccc}
\hline
\hline
Variant & RIDB & SE & \multicolumn{1}{l}{RaLS} \\ \hline
RIDB-Net & $\surd$ &  &  \\
RIDB-RaLS-Net & $\surd$ &  & $\surd$ \\
RIDB-SE-Net & $\surd$ & $\surd$ &  \\
RIDB-SE-RaLS-Net (SEGA-FURN) & \textbf{$\surd$} & \textbf{$\surd$} & $\surd$ \\ \hline \hline
\end{tabular}
%\end{center}
\end{table}

\begin{table}[t!]
\caption{Quantitative comparison of different variants on CelebA dataset for upscaling factor 4× and 8×.}
\small
% \begin{center}
% \centering
\setlength{\tabcolsep}{2.8mm}
\begin{tabular}{lll}
\hline
\hline
Ablation & \multicolumn{1}{c}{\begin{tabular}[c]{@{}c@{}}CelebA 4×\\ PSNR/SSIM\end{tabular}} & \multicolumn{1}{c}{\begin{tabular}[c]{@{}c@{}}CelebA 8×\\ PSNR/SSIM\end{tabular}} \\ \hline
(A) RIDB-Net & 28.64/0.8514 & 24.25/0.7177 \\
(B) RIDB-RaLS-Net & 28.71/0.8526 & 24.37/0.7218 \\
(C) RIDB-SE-Net & 29.60/0.8607 & 24.44/0.7181 \\
(D) \textbf{RIDB-SE-RaLS-Net} & \textbf{30.14/0.8682} & \textbf{25.21/0.7250} \\ \hline \hline

\end{tabular}
%\end{center}
\end{table}

\subsection{Ablation Study}
We further implemented ablation studies to investigate the performance of the proposed method. As shown in Table II, we list several variants based on different proposed components. First, among them, RIDB-Net is used as the baseline variant, which only contains a single component RIDB. Second, the RIDB-RaLS-Net is constructed by removing the Semantic Encoder (SE) from the SEGA-FURN. Next, RIDB-SE-Net means to remove RaLS loss of SEGA-FURN, and RIDB-SE-RaLS-Net equals to SEGA-FURN including all the three components. We provide the visual results of these variants in Fig. 5, and a quantitative comparison in Table III.

\textbf{Effect of RIDB} We compare the proposed RIDB with other feature extraction blocks, such as Residual Block (RB) from SRGAN \cite{SRGAN} and Residual in Residual Dense Block (RRDB) of ESRGAN \cite{ESRGAN_Wang}. As shown in Fig. 4 and Table I, the method using our generator employing RIDB outperforms SRGAN using RB and ESRGAN using RRDB in both qualitative and quantitative comparisons. The reason why our method outperforms the others is analyzed below. Different from RB, our RIDB introduces a densely connected structure that can combine different levels of features. In addition, different from RRDB, the proposed RIDB designs multi-level residual learning within each basic internal dense block, which is able to boost the flow of features through the generator and provide hierarchical features for the super-resolution process. Based on these observations and analysis, the effectiveness of the new generator using the proposed RIDB\cite{xiang2021end-to-end} is validated. 

\textbf{Effect of SE} The Ablation (A) and (C) performed by RIDB-Net and RIDB-SE-Net aim to illustrate the advantage of SE and also verify the effectiveness of the joint discriminator. The RIDB-SE-Net can obtain embedded semantics extracted by SE and further feed these semantics along with image data to the joint discriminator. In the training process, the embedded semantics is capable of providing useful semantic information for the joint discriminator. Such innovation can enhance the discriminative ability of the joint discriminator. Compared with RIDB-Net which does not employ SE and joint discriminator, the RIDB-SE-Net achieves significant improvements in terms of quantitative comparisons. Furthermore, as shown in Fig. 5, there is also a noticeable refinement in detailed texture. The enhanced performance can verify that the extracted embedded semantics has a superior impact on SR results and that the SE along with the joint discriminator plays a critical role in the proposed method.

\textbf{Effect of RaLS}
The ablation (A) and (B) are conducted to demonstrate the effect of RaLS loss. We replace the RaLS loss of RIDB-Net with the generic GAN loss, Binary Cross Entropy (BCE), and keep all the other components the same. As shown in Table III, it is obvious that once we remove the RaLS loss in RIDB-Net, the quantitative results are lower than RIDB-RaLS-Net which has RaLS loss. As expected, BCE used in RIDB-Net shows unrefined textures. In contrast, when RaLS is utilized in variant, the visual results are perceptually pleasant with more natural textures and edges. Thus, it can demonstrate that the RaLS loss is capable of greatly improving the performance of super-resolution.

\textbf{Final Effect} 
From the comparison between ablation (D), and other ablation variants, it is obvious that the large enhancement is noticeable by integrating all these three components together (RIDB, Semantic Encoder, and RaLS loss). Finally, we refer the RIDB-SE-RaLS-Net to SEGA-FURN which is the ultimate proposed method.

\section{Conclusion}
In this paper, we proposed a novel Semantic Encoder guided Generative Adversarial Face Ultra-resolution Network (SEGAFURN) to super-resolve a tiny LR unaligned face image to its HR version with multiple large ultra-upscaling factors (e.g., 4× and 8×). In SEGA-FURN, a novel Semantic Encoder, a novel Generator using Residual in Internal Dense Block, and a novel Joint Discriminator adopting RaLS loss, are proposed to solve the existing issues with other methods. The details of the three major components of the method are described in this paper, and their performance advantages over other state-of-the-art methods are analyzed. Extensive experiments demonstrate that SEGA-FURN is superior to state-of-the-art methods. Photo-realistic SR face images can be produced using the newly proposed SEGA-FURN method.

{\large
\bibliographystyle{IEEEtran}
%\bibliography{IEEEfull}
\bibliography{IEEEabrv}}

\newpage


\begin{IEEEbiography}[{\includegraphics[width=1in,height=1.25in,clip,keepaspectratio]{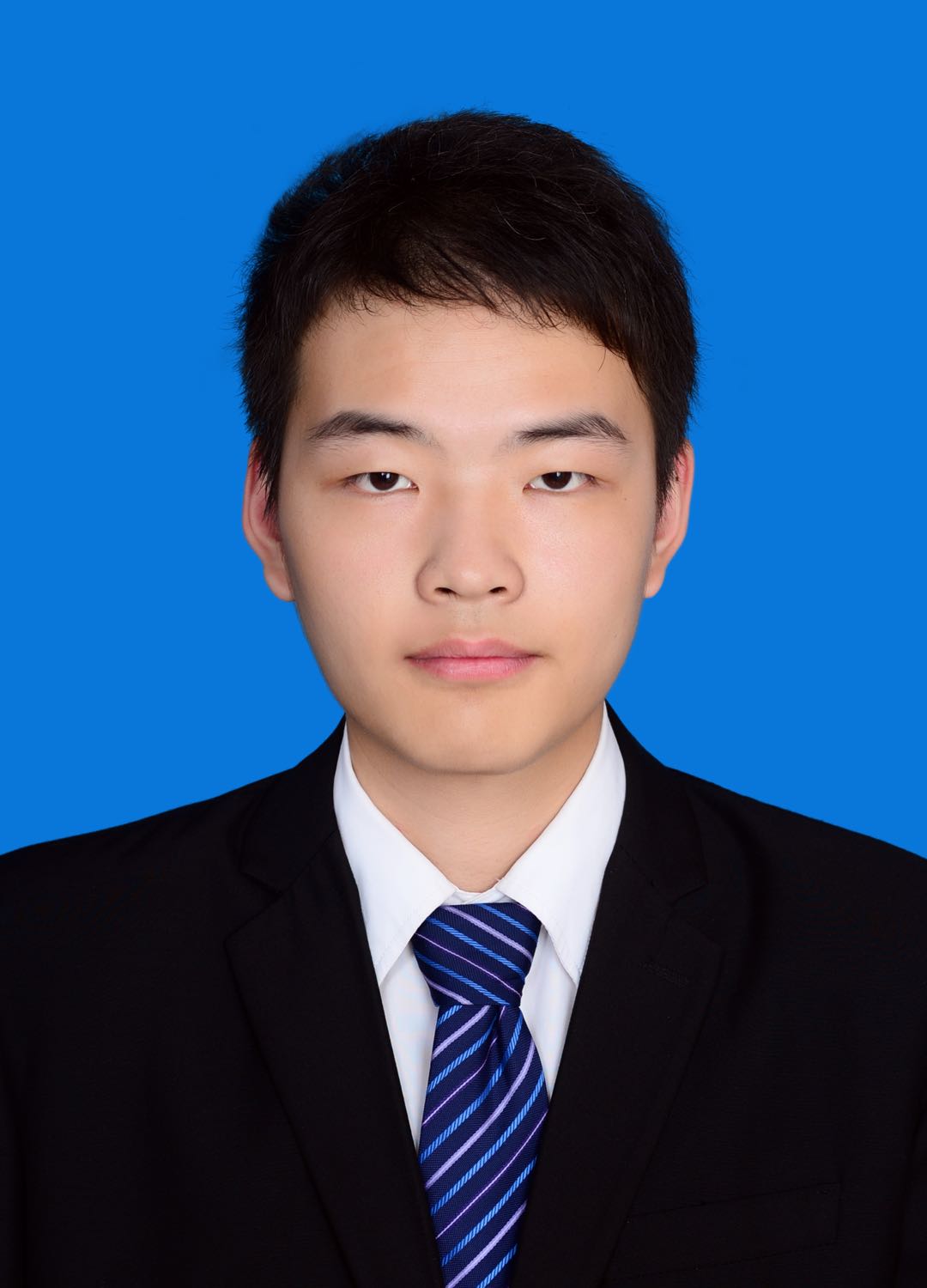}}]{Xiang Wang}
received the M.Sc. degree in computer science at Lakehead University, Thunder Bay, Canada. He worked as a research assistant at the Computer Vision and Machine Learning Laboratory in the Lakehead University for the research and design of computer vision and machine learning methods. His research interests include computer vision, image enhancement, image generation, and deep learning.
\end{IEEEbiography}

\begin{IEEEbiography}[{\includegraphics[width=1in,height=1.25in,clip,keepaspectratio]{./YY.pdf}}]{Yimin Yang (S’10-M’13-SM’19)} received his Ph.D. degrees in Pattern Recognition and Intelligent System from the College of Electrical and Information Engineering, Hunan University, China, in 2013.
He is currently an Assistant Professor at Department of electrical and computer engineering, Western University and he is also a faculty affiliate at the Vector Institute for Artificial Intelligence (Canada). From 2014 to 2018, he was a PostDoctoral Fellow with the Department of Electrical and Computer Engineering at the University of Windsor (Canada). He has authored or coauthored more than 50 refereed papers. His research interests are artificial neural networks, hybrid system approximation, and image feature selection.
Dr. Yang was the recipient of the Outstanding Ph.D. Thesis Award of Hunan Province, and the Outstanding Ph.D. Thesis Award Nominations of Chinese Association of Automation, China, in 2014 and 2015, respectively. He is an Associate Editor for the IEEE Transactions on Circuits and Systems for Video Technology, and the Neurocomputing. He has been serving as a Reviewer for many international journals of his research field and a Program Committee Member of some international conferences.
\end{IEEEbiography}

\begin{IEEEbiography}[{\includegraphics[width=1in,height=1.25in,clip,keepaspectratio]{./pang.pdf}}]{Qixiang Pang} is currently a professor at the Okanagan College, Canada. Previously, he worked with Motorola, TELUS, General Dynamics, SED, the University of British Columbia, and, the Chinese University of Hong Kong. He received his Ph.D. degree from Beijing University of Posts and Telecommunications, China. He did his postdoctoral research at the University of British Columbia, Canada. His research interests lie in Cloud/Edge Computing, IoT, 6G, and AI. He has been serving on the Technical Program Committee of many international conferences.
\end{IEEEbiography}

\begin{IEEEbiography}[{\includegraphics[width=1in,height=1.25in,clip,keepaspectratio]{./luxiao.pdf}}]{Xiao Lu} received the Ph.D. degree in pattern recognition and intelligent system from the College of Electrical and Information Engineering, Hunan University, Changsha, China, in 2015. She is currently an Assistant Professor with the College of Engineering and Design, Hunan Normal University, Changsha, China. Her current research interests include computer vision, machine learning, and robotics.
\end{IEEEbiography}
\begin{IEEEbiography}[{\includegraphics[width=1in,height=1.25in,clip,keepaspectratio]{./liuyu.pdf}}]{Yu Liu} received the B.Eng in electronic engineering from Tianjin University, Tianjin, China in 1998. During 1998-2000, he was an electronic engineer of Nantian Electronics Information Corp, Shenzhen, China. In 2000, he backed to Tianjin University and received M.Eng in information and communication engineering and the Ph.D. degree in signal and information processing from the same University in 2002 and 2005 respectively. Currently, he is a professor with School of Microelectronics, Tianjin University. During 2011-2012, Dr. Liu was a visiting fellow with the Department of Electrical Engineering, Princeton University, Princeton, NJ, U.S. His research interests include signal/video processing, medical signal processing, multimedia systems, compressed sensing, indoor positioning system and machine learning.
\end{IEEEbiography}

\begin{IEEEbiography}[{\includegraphics[width=1in,height=1.25in,clip,keepaspectratio]{./shandu.pdf}}]{Shan Du (S’05-M’09-SM’16)} received the Ph.D. degree in Electrical and Computer Engineering from the University of British Columbia, Vancouver, BC, Canada in 2009. She is currently an assistant professor with the Department of Computer Science, Mathematics, Physics \& Statistics, the University of British Columbia (Okanagan Campus). Before joining UBC, she was working as an assistant professor with the Department of Computer Science, Lakehead University, Canada and as a Research Scientist/Software Engineer with IntelliView Technologies Inc., Canada. Shan has more than 15 years research and development experience on image/video processing, image/video analytics, pattern recognition, computer vision and machine learning. Shan was recipient of many awards and grants, including NSERC-IRDF, NSERC-CGS D, AITF Industry r\&D Associates Grant, and ICASSP Best Paper Award. Shan is a senior member of IEEE, IEEE Signal Processing Society and IEEE Circuits and Systems Society. She is serving as an Associate Editor of IEEE Trans. on Circuits and Systems for Video Technology and IEEE Canadian Journal of Electrical and Computer Engineering, Area Chair of ICIP 2021, and served as Area Chair and Session Chair of ICIP 2019, TPC member and reviewer for many international journals and conferences.
\end{IEEEbiography}

\end{document}